\documentclass{article}
\usepackage{PRIMEarxiv}

\usepackage{amsmath,amsfonts}
\usepackage{algorithmic}
\usepackage{algorithm}
\usepackage{array}
\usepackage[caption=false,font=normalsize,labelfont=sf,textfont=sf]{subfig}
\usepackage{textcomp}
\usepackage{stfloats}
\usepackage{url}
\usepackage{verbatim}
\usepackage{graphicx}
\usepackage{booktabs}
\usepackage{float}
\usepackage[flushleft]{threeparttable}
\usepackage{multirow}
\usepackage{setspace}
\usepackage{makecell} 
\usepackage{soul, color}
\usepackage{flushend}
\usepackage{ulem}
\usepackage{enumitem}
\usepackage{hyperref}
\hypersetup{
        colorlinks=true,
        linkcolor=blue,
        filecolor=green,      
        urlcolor=blue, %cyan
        citecolor=red,
}

\title{CS-SHAP: Extending SHAP to Cyclic-Spectral Domain for Better Interpretability of Intelligent Fault Diagnosis}

\author{
  Qian Chen \\
  Shanghai Jiao Tong University \\
  \texttt{chenqian2020@sjtu.edu.cn} \\
     \And
  Xingjian Dong \thanks{Corresponding Author.} \\
  Shanghai Jiao Tong University \\
  \texttt{donxij@sjtu.edu.cn} \\
     \And
Kui Hu\\
  Shanghai Jiao Tong University \\
  \texttt{hu\_kui@sjtu.edu.cn} \\   
  \And
  Kangkang Chen\\
  Shanghai Jiao Tong University \\
  \texttt{chenkk1220@sjtu.edu.cn} \\   
    \And
    Zhike Peng\\
  Shanghai Jiao Tong University \\
  \texttt{z.peng@sjtu.edu.cn} \\
     \And
     Guang Meng\\
  Shanghai Jiao Tong University \\
  \texttt{gmeng@sjtu.edu.cn} \\
}

\begin{document}
\maketitle

\begin{abstract}
    Neural networks (NNs), with their powerful nonlinear mapping and end-to-end capabilities, are widely applied in mechanical intelligent fault diagnosis (IFD). However, as typical black-box models, they pose challenges in understanding their decision basis and logic, limiting their deployment in high-reliability scenarios. Hence, various methods have been proposed to enhance the interpretability of IFD. Among these, post-hoc approaches can provide explanations without changing model architecture, preserving its flexibility and scalability. However, existing post-hoc methods often suffer from limitations in explanation forms. They either require preprocessing that disrupts the end-to-end nature or overlook fault mechanisms, leading to suboptimal explanations. To address these issues, we derived the cyclic-spectral (CS) transform and proposed the CS-SHAP by extending Shapley additive explanations (SHAP) to the CS domain. CS-SHAP can evaluate contributions from both carrier and modulation frequencies, aligning more closely with fault mechanisms and delivering clearer and more accurate explanations. Three datasets are utilized to validate the superior interpretability of CS-SHAP, ensuring its correctness, reproducibility, and practical performance. With open-source code and outstanding interpretability, CS-SHAP has the potential to be widely adopted and become the post-hoc interpretability benchmark in IFD, even in other classification tasks. The code is available on \textit{\url{https://github.com/ChenQian0618/CS-SHAP}}.
\end{abstract}


\section{Introduction}

Beyond design and manufacturing, maintenance is also important in the lifecycle of mechanical equipment~\cite{leiMachineryHealthPrognostics2018}. Mechanical fault diagnosis (FD) plays a vital role by identifying fault types or locations after their occurrence, enabling targeted maintenance~\cite{fengReviewVibrationbasedGear2023}. It not only significantly reduces costs but also minimizes equipment downtime. Consequently, FD is widely utilized in industrial production and equipment maintenance~\cite{gangsarSignalBasedCondition2020}.

With the growing complexity of mechanical equipment and advancements in sensor technology, the deployment of numerous sensors generates vast amounts of operational data, marking the advent of big data era~\cite{leiApplicationsMachineLearning2020}. Traditional fault diagnosis methods, which depend on prior knowledge and manual processing, are inadequate for the complex diagnostic scenarios posed by big data. In contrast, intelligent fault diagnosis (IFD) utilizes abundant data without relying on prior knowledge, effectively meeting the requirements of big data~\cite{qianFederatedTransferLearning2025}. %~\cite{panGenerativeAdversarialNetwork2022,lvAttentionMechanismIntelligent2022a,qianFederatedTransferLearning2025}

IFD, leveraging the nonlinear mapping of neural networks (NNs), offers an end-to-end feasible solution. IFD has developed into a mature and thriving research domain, with numerous advanced NNs being progressively applied to diagnose mechanical components such as bearings~\cite{hakimSystematicReviewRolling2023}, gears~\cite{liSpectralSelffocusingFault2023}, and pumps~\cite{xuPhysicsConstraintVariationalNeural2024}. Although significant progress has been made in areas such as diagnostic accuracy, generalization capability~\cite{yangLabelRecoveryTrajectory2024,liFederatedTransferLearning2023a}, information fusion~\cite{xuCFCNNNovelConvolutional2023}, and few-shot learning~\cite{shaoFewShotCrossDomainFault2024,liLabelguidedcontrastivelearning2025}, the interpretability of IFD remains underexplored. % liuOptimalSubdomainGeneralizationMethod2024

Neural networks, as the foundation of IFD, involve multiple nonlinear mappings and are typical \textit{black boxes}. It is challenging to understand their reasoning basis, logical processes, and applicable scopes~\cite{ivanovsPerturbationbasedMethodsExplaining2021}. However, interpretability is critical for the practical application of IFD~\cite{zhaoChallengesOpportunitiesAIEnabled2021}. From the user's perspective, non-interpretable IFD struggle to gain trust, often necessitating cross-validation, which increases maintenance costs. From the developer's perspective, the lack of interpretability hinders the ability to identify errors and optimize models scientifically. From the application perspective, real-world scenarios are more complex and unpredictable than training data, making it difficult for non-interpretable IFD systems to ensure reliable performance under actual operating conditions. Therefore, the interpretability research of IFD has substantial scientific and industrial significance.

Interpretability refers to \textit{the ability to provide explanations in understandable terms to a human}~\cite{zhangSurveyNeuralNetwork2021}. Existing interpreting methods are generally categorized into two types: ante-hoc (active) and post-hoc (passive).

% Post-hoc interpretability methods, such as SHAP, LIME, and DeepLIFT.

Ante-hoc interpretability requires modifications before training, such as changing architectures or optimization processes. For instance, An~et~al.~\cite{anAdversarialAlgorithmUnrolling2024} employed algorithm unrolling to design interpretable NN structures that capture fault characteristics through learned dictionaries. Li~et~al.~\cite{liVariationalAttentionBasedInterpretable2024} enhanced traditional transformers by leveraging the multi-head attention mechanism to explain segment contributions. Wang~et~al.~\cite{wangFullyInterpretableNeural2022} integrated traditional feature extraction techniques to design extreme learning machines (ELMs) capable of localizing resonance frequency bands. However, while these modifications provide NNs with distinct interpretability, they also constrain model structures, thereby reducing the flexibility and scalability. Moreover, ante-hoc interpretability can sometimes compromise performance. Despite claims of high accuracy in various studies~\cite{wangInterpretableConvolutionalNeural2023,chenInterpretingWhatTypical2024,chenTFNInterpretableNeural2024}, the constrained architectures inherently limit the potential for further optimization.

In contrast, post-hoc interpretability focuses on explaining pre-trained networks without requiring modifications before training. This approach avoids introducing additional constraints, ensuring models remain flexible and scalable without sacrificing performance. Current post-hoc interpretability in IFD primarily focuses on attribution, which aims to reveal the basis of decisions by \textit{assigning credit (or blame) to the input features}~\cite{zhangSurveyNeuralNetwork2021} (e.g., feature importance). Typical attribution methods include class activation mapping (CAM), layer-wise relevance propagation (LRP), and \underline{SH}apley \underline{A}dditive ex\underline{P}lanations (SHAP), etc. Inspired by attribution methods in computer vision, some researchers have extended them to IFD. For example, Wu~et~al.~\cite{wuHybridClassificationAutoencoder2021} and Grezmak~et~al.~\cite{grezmakInterpretableConvolutionalNeural2020} applied time-frequency transform to convert vibration signals into 2D images, then used existing Grad-CAM and LRP to compute contributions, respectively. However, this approach achieved suboptimal interpretability and is not end-to-end. To achive end-to-end interpretability, Li~et~al.~\cite{liWhiteningNetGeneralizedNetwork2021} use the integrated gradient (IG) method to effectively capture the characteristic frequency through \textit{Fourier} transform, and Li~et~al.~\cite{liMultilayerGradCAMEffective2023} made further improvement by combining Grad-CAM with correlation-based weighting. Despite these advancements, the time-domain explanations remain insufficiently intuitive and often require \textit{Fourier} transform for clearer analysis.

The core challenge in post-hoc interpretability for IFD is the form of explanation. End-to-end models typically process time-domain signals as input, but these signals often fail to directly convey mechanistic insights~\cite{zhangVibrationFeatureExtraction2022}. Consequently, transforming signals from the time domain to more clear domains is a well-established practice in traditional FD~\cite{yanWaveletTransformRotary2023}. Likewise, extending the explanation to these clearer domains offers a feasible approach for enhancing the explainability of IFD.

\begin{figure}[t]
        \centering
        \includegraphics[width=\textwidth]{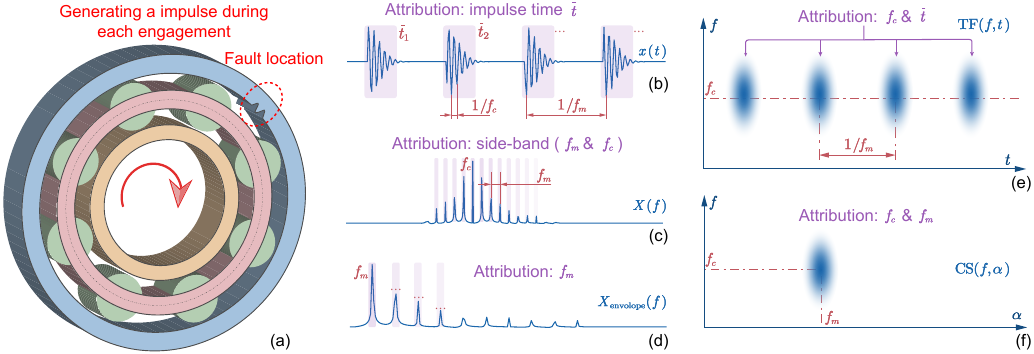}
        \caption{The schematic diagram of the relationship between rotating machinery fault and vibration signal, and the characteristics and attribution results of vibration signal in different domains. 
        Here, $f_c$ represents the carrier frequency, corresponding to the fault resonance frequency, while $f_m$ denotes the modulation frequency, corresponding to the fault excitation frequency. (a) A bearing with outer race fault, will generate impulses during each engagement periodically. (b) Time domain. (c) Frequency domain. (d) Envelope domain. (e) Time-frequency domain. (f) Cyclic-spectral (CS) domain.}
        \label{fig:intro}
\end{figure}

To facilitate understanding, Fig.~\ref{fig:intro} illustrates fault representations across different domains. When a fault occurs, as shown in Fig.~\ref{fig:intro}(a), the fault component generates periodic impulse responses during rotation, producing the time-domain signal depicted in Fig.~\ref{fig:intro}(b). However, the time domain is prone to noise interference, and attribution methods often only capture impulse timings. Therefore, domain transformation can provide clearer fault representations. In the frequency domain shown in Fig.~\ref{fig:intro}(c), fault components appear as sidebands, enabling the explanation of the carrier frequency $f_c$ and indirectly the modulation frequency $f_m$. In the envelope domain shown in Fig.~\ref{fig:intro}(d), fault components manifest as modulation frequencies $f_m$ and their harmonics, enabling attribution results to locate $f_m$. In the time-frequency domain shown in Fig.~\ref{fig:intro}(e), faults are represented as periodic impulses, allowing attribution methods to simultaneously reveal the carrier frequency $f_c$ and impulse timings.

Based on this, some researchers have optimized post-hoc explanation forms by incorporating domain transformations~\cite{borghesaniFourierbasedExplanation1DCNNs2023,wuInterpretableMultiplicationConvolutionSparse2023}. For example, Gwak~et~al.~\cite{gwakRobustExplainableFault2023} perturbed samples in the frequency domain to identify critical frequencies and decision boundaries of end-to-end models. Herwig~et~al.~\cite{herwigExplainingDeepNeural2023} extended SHAP to the frequency and time-frequency domains, effectively revealing the contributions of different frequencies $f_c$ from time-domain samples. Going further, Decker~et~al.~\cite{deckerDoesYourModel2023} applied SHAP to the envelope domain, providing attribution for modulation frequencies $f_m$ which is better suited for bearing faults.

However, these methods only analyze faults from either the carrier frequency $f_c$ or modulation frequency $f_m$ perspective alone, which may result in partial or even misleading explanations. Actually, many faults may share similar $f_c$ or $f_m$, making it challenging for these methods to distinguish such close fault components effectively. In contrast, as shown in Fig.~\ref{fig:intro}(f), the cyclic-spectral (CS) domain~\cite{randallVibrationbasedConditionMonitoring2022} simultaneously represents fault components through both $f_c$ and $f_m$. This dual-dimensional analysis provides superior fault discrimination capabilities, enabling more comprehensive and accurate explanations.

Thus, we propose CS-SHAP by extending SHAP attribution to the more informative CS domain, achieving superior interpretability. Specifically, we first innovatively derive the CS transform $\mathcal{Z}$ and its corresponding inverse transform $\mathcal{Z}^{-1}$ for deterministic signals, as traditional CS analysis is designed for stochastic signals. Then, leveraging SHAP, we can attribute explanations in the CS domain without altering the end-to-end architecture. The proposed CS-SHAP can attribute contributions simultaneously from both carrier $f_c$ and modulation $f_m$ frequencies, aligning closely with fault mechanisms and effectively distinguishing close fault components. This results in clearer and more precise explanations. As a post-hoc interpretability algorithm, CS-SHAP is model-agnostic. With superior interpretability and open-source code, CS-SHAP is poised to become a universal post-hoc interpretability benchmark in the IFD domain. Our contributions can be summarized as follows:

\begin{enumerate}
        \item The CS transform and its inverse are innovatively derived from correlation analysis, forming the foundation for the SHAP extension.
        \item CS-SHAP attributes contributions from both carrier and modulation frequencies while preserving the end-to-end structure, aligning with fault mechanisms to provide clearer and more accurate post-hoc explanations.
        \item Extensive evaluations on three datasets, combined with open-source code availability, establish CS-SHAP as a superior and reliable interpretability benchmark, highlighting its potential for widespread adoption in IFD.
\end{enumerate}

\section{Preliminary}

\subsection{SHAP}

SHAP (\underline{SH}apley \underline{A}dditive ex\underline{P}lanations)~\cite{lundbergUnifiedApproachInterpreting2017} is a game-theory-inspired approach designed to explain machine learning model predictions. It quantifies the contribution of each feature in a sample to the model’s prediction, i.e., attribution. To understand SHAP, it is essential to first introduce the Shapley value~\cite{shapleyValueNpersonGames1953}, which is a mathematical method for determining a fair and efficient resource allocation strategy within a group. Its applications include profit sharing among stakeholders, cost distribution among collaborators, etc.

Denoting the players as $p$, the set of players as $S$, and the value of the set as $v:S\rightarrow \mathbb{R}$. The marginal contribution $\Delta v (p,S)$, which measures the effect of player $p$ joining the set $S$, can be expressed as:
\begin{equation}
        \Delta v (p,S) = v ( S \cup\{p \} )-v (S) .
        \label{eq:SHAP_delta_v}
\end{equation}
Therefore, the Shapley value $\psi(v,U)_{i}$, representing the expected marginal contribution of player $p_i$ across all possible subsets $S$ of the universe set $U=\{p_1,p_2,\cdots,p_n\}$, can be expressed as:
\begin{equation}
        \psi(v,U)_{i}=\sum_{S \subseteq U \setminus\{p_i \}} \frac{s! ( n-s-1 )!} {n!} \cdot \Delta v  (p_i,S)
        \label{eq:SHAP_Shapley}
\end{equation}
where, $s$ represents the number of elements of $S$, and the weight ${s! ( n-s-1 )!} \,/\,{n!}$ represents the probability of player $p_i$ joining the subset $S$.

The Shapley value effectively measures the contribution of each set member but is not directly applicable to machine learning. In Shapley value, the input dimension (i.e., the number of set members) of the value function $v$ can be different, whereas the input dimension for machine learning is fixed.

To address this issue, SHAP treats all the feature dimensions as the universe set $U=\{1,2,\cdots,d\}$, with $d$ being the number of feature dimensions. It then treats the features within the subset as fixed values and features outside the subset as variables. The value function is then constructed by calculating the expected output over the distribution of these variables. For a given subset $S$, features within $S$ retain their fixed values from the input sample $\tilde x$, while features outside $S$ are sampled from the data distribution $X$. The resulting samples can be expressed as:
\begin{equation}
        \tilde x_{S,i} =\left\{ 
    \begin{array}{ll}
        \tilde x_i \, \textrm{(constant)}, &\textrm{if}\; i \, \in\, S\\
        X_i \, \textrm{(variable)}, \quad &\textrm{if}\; i \, \notin\, S.
    \end{array}
\right.
\end{equation}
Denoting the model as $\mathcal{M}:\mathbb{R}^d \rightarrow \mathbb{R}$, the value function $v_{\mathcal{M}, x,\tilde x} ( S )$ in SHAP can be defined as:
\begin{equation}
        \begin{split}
        v_{\mathcal{M}, x, \tilde x} ( S )&= \mathbb{E} [\mathcal{M} ( \tilde x_S)]-\mathbb{E} [\mathcal{M} ( x )],\\
         &= \int\! \mathcal{M}(\tilde x_S) \,\mathrm{d} \mathbb{P}_{X} - \int\! \mathcal{M}(x) \,\mathrm{d} \mathbb{P}_{X}.
        \end{split}
\end{equation}
By substituting $v_{\mathcal{M},x,\tilde x} ( S )$ into \eqref{eq:SHAP_delta_v} and \eqref{eq:SHAP_Shapley}, the SHAP result $\psi_{\mathcal{M},X}(\tilde x) $ can be computed as:
\begin{equation}
        \begin{split}
       &\psi_{\mathcal{M},X}(\tilde x) _i \!= \\
       &\sum_{S \subseteq U \setminus\{i \}} \! \frac{s! ( d-s-1 )!} {d!} \left( \mathbb{E} [\mathcal{M} ( \tilde x_{S \cup \{i\}})]-\mathbb{E} [\mathcal{M} ( \tilde x_S)]\right).
        \end{split}
        \label{eq:SHAP_SHAP}
\end{equation}

The computation of SHAP requires enumerating feature subsets $S$ from $U$ and estimating expectations $\mathbb{E} [f(x)]$ over the data distribution, making the implementation quite challenging. To address this, Lundberg~et~al.~\cite{lundbergUnifiedApproachInterpreting2017} developed an open-source Python library~\cite{Shap2024} that simplifies and facilitates the application of SHAP.

\subsection{Cyclic-spectral correlation}
\label{subsection:CS}

Cyclic-spectral correlation (CSC) is an effective signal processing technique for analyzing cyclostationary signals~\cite{randallVibrationbasedConditionMonitoring2022}. Unlike stationary signals, which have constant statistical characteristics, a N-order cyclostationary signal is one whose N-th order statistics are periodic over time.

For instance, a second-order cyclostationary signal has a periodic autocorrelation function, such as white noise modulated by a periodic signal. These signals typically represent system responses to periodic excitations. As illustrated in Fig.~\ref{fig:intro}(a), rotating machinery induces periodic contact of fault components during rotation, thereby exciting the system response periodically. The resulting vibration signal, depicted in Fig.~\ref{fig:intro}(b), exemplifies a typical second-order cyclostationary signal. This cyclostationary signal has two types of information: the system response frequency ($f_m$) and the periodic excitation frequency ($f_c$). While traditional spectral analysis is inadequate for such signal, CSC can effectively reveal its hidden features.

For a cyclostationary signal $x(t)$, its second-order cyclic stationary moments $R_{x}(\tau,t)$ can be computed using a two-dimensional autocorrelation function:
\begin{equation}
        R_{x}(\tau,t)=E[x(t-\tau/2)x^*(t+\tau/2)]
        \label{eq:CSC_ACF}
\end{equation}
where, $*$ denotes the conjugate operator, $\tau$ represents the time-delay, and $E$ stands for the statistical expectation. This function represents the time-domain characteristics of $x(t)$ at a specific time $t$. The $\tau$-axis captures the system response information, while the $t$-axis reflects the excitation information.

The cyclic autocorrelation function (CAF) $R_{x}(\tau,\alpha)$ is derived by applying \textit{Fourier} transform (FT) to the time axis $t$ of the second-order cyclostationary moments:
\begin{equation}
        \begin{split}
                R_{x}&(\tau,\alpha) = \int R_x(\tau,t) e^{-i2\pi \alpha t} \,{\mathrm d}t \\
                % &\approx \lim_{T\rightarrow \infty} \frac{1}{T} \int^{T/2}_{-T/2} x(t-\tau/2)x^*(t+\tau/2) e^{-i2\pi \alpha t}\,{\mathrm d}t
        \end{split}
\end{equation}
where, $\alpha$ denotes the cyclic frequency.
% , and the approximation holds true when the signal is cycloergodic~\cite{CyclicAutocorrelationFunction}.
 $R_{x}(\tau,\alpha)$ provides the time-domain characterization of $x(t)$ at different cyclic frequencies $\alpha$, effectively exposing the frequency information of the excitation.

By performing FT on the time-delay axis $\tau$ of the CAF $R_{x}(\tau,\alpha)$, the CSC can be obtained:
\begin{equation}
        \begin{split}
                S_x{(f,\alpha)} &=\int R_x(\tau,\alpha) e^{-i2\pi f \tau} \,{\mathrm d}\tau\\
                &= \int \int R_x(t,\tau)e^{-i2\pi (f \tau + \alpha (t) } \,{\mathrm d}t \,{\mathrm d}\tau
        \end{split}
        \label{eq:CSC_CSC}
\end{equation}
where, $f$ represents the spectral frequency. The CSC can be regarded as the two-dimensional \textit{Fourier} transform of the second-order cyclostationary moments $R_{x}(\tau,t)$ along the time $t$ and time-delay $\tau$ axes, yielding a function of spectral frequency $f$ and cyclic frequency $\alpha$.

Unlike traditional spectral analysis, CSC introduces an additional dimension for cyclic frequency, allowing for the simultaneous identification of carrier and modulation components in fault signals. Specifically, the spectral frequency $f$ corresponds to the carrier component $f_c$, while the cyclic frequency $\alpha$ represents the modulation component $f_m$. This method has widely applied in stochastic signal analysis since last century~\cite{spoonerCumulantTheoryCyclostationary1994,gardnerCumulantTheoryCyclostationary1994}, but also adopted for data preprocessing in IFD recently~\cite{chenDeepLearningMethod2020}.

\section{Method}
\subsection{Cyclic-spectral transform}
For end-to-end IFD models, existing SHAP methods can attribute contributions in the time domain or extend to frequency, time-frequency, and envelope domains. However, these approaches often fall short of comprehensively revealing fault components. The CSC method introduced in Section~\ref{subsection:CS}, has the ability to simultaneously uncover carrier and modulation components, and inspires our motivation of extending SHAP to the CS domain.

However, CSC is inherently designed for stochastic signals and is not directly applicable to the deterministic signals used in IFD models. The key to applying CSC to deterministic signals lies in estimating the second-order cyclostationary moment $R_{x}(\tau,t)$ as mentioned in~\eqref{eq:CSC_ACF}. For second-order stationary signals, increasing the time delay $\tau$ would cause $R_{x}(\tau,t)$ to decay to zero. Essentially, it represents the autocorrelation function (ACF) of the signal $x(t)$ at a specific local time $t$. Therefore, we can apply a window $h(t)$ to the deterministic signal $x(t)$ and compute its ACF as an approximation $R'_{x}(\tau,t)$. The windowed signal $\hat x(t',t)$ can be expressed as:
\begin{equation}
        \hat x(t',t)=x(t'-t)h(t')
\end{equation}
where, $t'$ represents the time-axis, and $t$ denotes the center of the window. Then, $R'_x(\tau,t)$ can be derived from the ACF $\mathcal{R}(\cdot)$ of $\hat x(t',t)$ along axis $t'$:
\begin{equation}
        \begin{split}
                R'_x(\tau,t)&=\mathcal{R}_{t'\rightarrow\tau}(\hat x(t',t))\\
                &=\int \hat x(t'-\tau/2,t)\cdot \hat x^*(t'+\tau/2,t) \,\mathrm{d}t'.
        \end{split}
\end{equation}

By substituting the approximate moment $R'_x(\tau,t)$ into \eqref{eq:CSC_CSC}, the cyclic-spectral (CS) representation $\mathrm{CS}_x(f,\alpha)$ of the input signal $x(t)$ can be expressed as:
\begin{equation}
        \begin{split}
                \mathrm{CS}_x(f,\alpha)\!&=\int \int R_x(\tau,t)e^{-i2\pi (f \tau + \alpha t) } \,{\mathrm d}\tau \,{\mathrm d}t\\
                &= \int \Big[ \int \mathcal{R}_{t'\rightarrow\tau} \big(\hat x(t',t)\big) e^{-i2\pi f \tau} {\mathrm d}\tau \Big] \cdot e^{-i2\pi \alpha t} \,{\mathrm d}t\\
                &= \int  \mathcal{F}_{\tau\rightarrow f}\left( \mathcal{R}_{t'\rightarrow\tau}\big(\hat x(t',t)\big)\right) \cdot e^{-i2\pi \alpha t} \,{\mathrm d}t
        \end{split}
        \label{eq:CS_CSv1}
\end{equation}
where, $\mathcal{F}(\cdot)$ denotes the \textit{Fourier} transform (FT), and the FT of ACF yields the power spectral density (PSD). The relationship between the PSD and the FT is given by:
\begin{equation}
        \mathcal{F}_{\tau\rightarrow f}\Big( \mathcal{R}_{t'\rightarrow\tau}\big( x(t') \big)\Big) =\Big| \mathcal{F}_{t\rightarrow f}\big( x(t') \big) \Big|^2
        \label{eq:CS_PSD}
\end{equation}
Based on \eqref{eq:CS_CSv1} and \eqref{eq:CS_PSD}, it can be deduced that:
\begin{equation}
        \begin{split}
                \mathrm{CS}_x(f,\alpha)
                &=\int \left| \mathcal{F}_{t'\rightarrow f}\left(\hat x(t',t) \right) \right|^2 \cdot e^{-i2\pi \alpha t} \,{\mathrm d}t\\
                &=\mathcal{F}_{t\rightarrow \alpha} \left[ \left| \int x(t'-t)h(t') e^{-i2\pi f t'} {\mathrm d}t' \right|^2 \right]\\
               % &= \int \left| \mathrm{STFT}_x(f,t) \right|^2 \cdot e^{-i2\pi \alpha t} \,{\mathrm d}t\\
                &=\mathcal{F}_{t\rightarrow \alpha} \left[ \left| \mathrm{STFT}_x(f,t) \right|^2 \right].
        \end{split}
\end{equation}
where, $\mathrm{STFT}_x(f,t)$ denote the short-time \textit{Fourier} transform of signal $x(t)$.
In conclusion, by applying FT to the time-axis $t$ of $\left| \mathrm{STFT}_x(f,t) \right|^2$ of a deterministic signal $x(t)$, we can obtain its representation $\mathrm{CS}_x(f,\alpha)$ in the CS domain. While this approach was implicitly conducted in Literature~\cite{liuNTScatNetInterpretableConvolutional2022}, to the best of our knowledge, we are the first to rigorously derive and formally prove it.

The above work presents the process of transforming a deterministic signal into the CS domain, named as the CS transform $\mathcal{Z}$. In addition to second-order cyclostationary signals, $\mathcal{Z}$ is also suitable for common sine signals, effectively revealing their characteristic frequencies. Specifically, for the sine signal $x_1(t)$:
\begin{equation}
        x_1(t) = A_1 \sin(2\mathrm{\pi}f_1t+\phi_1)
\end{equation}
where, $A_1$, $f_1$, $\phi_1$ represent the amplitude, fault frequency, and phase of $x_1(t)$, respectively. Then, the STFT energy of $x_1(t)$ is given by:
\begin{equation}
                \left| \mathrm{STFT}_x(f,t) \right|^2 = \left\{ \begin{array}{ll}
                        K, \quad &\textrm{if} \; f\!=\! \pm f_1\\
                0, \quad &\textrm{else}
            \end{array} \right., \quad
            K=\frac{A_1}{2}\cdot  \left| \int h(t') \, {\mathrm d}t' \right|^2.
\end{equation}

Furthermore, the CS representation of the sine signal $x_1(t)$ can be expressed as:
\begin{equation}
        \begin{split}
                \mathrm{CS}_{x_1}(f,\alpha) \!&=\! \left\{ \begin{array}{ll}
                        K T, \quad &\textrm{if} \; f\!=\!\pm f_1,\alpha\!=\!0\\
                0, \quad &\textrm{else}
            \end{array} \right. 
        \end{split}
\end{equation}
where, $T$ represents the signal length of $x_1(t)$. In summary, a sinc signal can be regarded as a second-order cyclostationary signal with cyclic frequency $\alpha=0$. Its CS representation will appear at $f\!=\!\pm f_1,\alpha\!=\!0$, indicating that the proposed CS transform $\mathcal{Z}$ remains applicable to common sinc signals.

\begin{figure}[htbp]
        \centering
        \includegraphics[width=\textwidth]{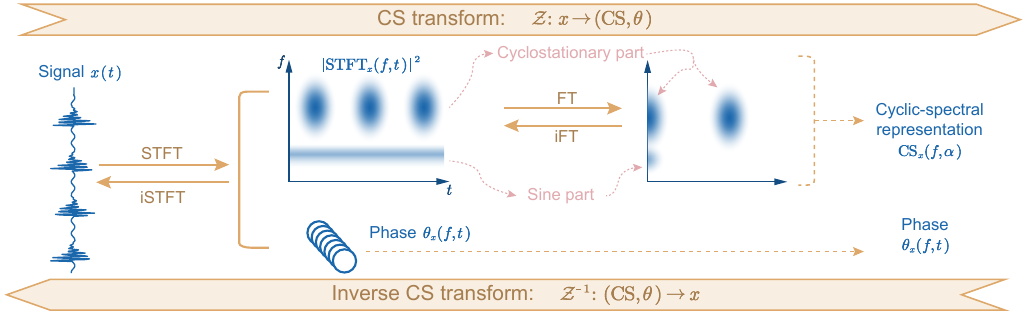}
        \caption{The processes of the CS transform and inverse CS transform.}
        \label{fig:3-transform}
\end{figure}

To integrate this approach with SHAP, the inverse CS transform $\mathcal{Z}^{-1}$ must also be established. The processes of $\mathcal{Z}$ and $\mathcal{Z}^{-1}$ are illustrated in Fig.~\ref{fig:3-transform}, and the phase information $\theta_x(f,t)$ is additionally preserved to ensure accurate reconstruction in the inverse CS transform, avoiding the information loss caused by the modulus operation.

Given an input signal $x(t)$, the CS transform $\mathcal{Z} :\;x\rightarrow (\mathrm{CS},\theta )$ can be expressed as:
\begin{equation}
        \begin{split}
                \mathrm{STFT}_x(f,t)&=\int x(t'-t)h(t') e^{-i2\pi f t'} {\mathrm d}t',\\
                \mathrm{CS}_x(f,\alpha)&=\int \left| \mathrm{STFT}_x(f,t) \right|^2 \cdot e^{-i2\pi \alpha t} \,{\mathrm d}t,\\
                \theta_x(f,t)&=\mathrm{Angle}(\mathrm{STFT}_x(f,t))\\
        \end{split}
        \label{eq:CS_Z}
\end{equation}
where, $\mathrm{Angle}(\cdot)$ represents the function used to extract the phase of a complex value.

Given the CS representation $\mathrm{CS}_x(f,\alpha)$ and its phase information $\theta_x(f,t)$, the inverse CS transform $\mathcal{Z} ^{-1}:\;(\mathrm{CS},\theta )\rightarrow x$ can be expressed as
\begin{equation}
        \begin{split}
                \mathrm{STFT}_x(f,t) &=\sqrt {\int \mathrm{CS}_x(f,\alpha) \cdot e^{i2\pi \alpha t} \,{\mathrm d}t} \cdot (\cos\theta+i\sin\theta),\\
               x(t)&=\mathrm{iSTFT}\left( \mathrm{STFT}_x(f,t) \right)
        \end{split}
        \label{eq:CS_Z_1}
\end{equation}
where, $\mathrm{iSTFT}(\cdot)$ is the inverse short-time \textit{Fourier} transform.

With the CS transform $\mathcal{Z}$ and inverse CS transform $\mathcal{Z}^{-1}$, we can achieve mutual transformation between the time domain and the CS domain, which will facilitate the subsequent implementation of CS-SHAP.

\subsection{CS-SHAP}

\begin{figure}[htbp]
        \centering
        \includegraphics[width=\textwidth]{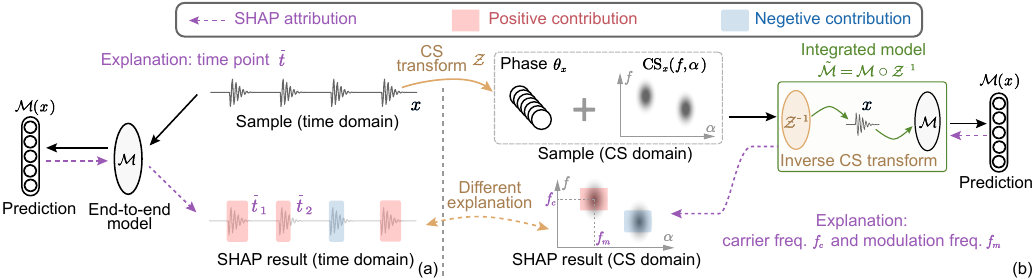}
        \caption{The illustration of SHAP attribution methods in time-domain and CS-domain. (a) The traditonal time-domain SHAP, whose result is the components with different timings. (b) The proposed CS-SHAP, whose result is the components with different spectral frequency and cyclic frequency.}
        \label{fig:method}
\end{figure}

By leveraging the CS transform and inverse CS transform shown in \eqref{eq:CS_Z} and \eqref{eq:CS_Z_1}, traditional time-domain SHAP can be extended to the CS domain, as illustrated in Fig.~\ref{fig:method}.

Traditional time-domain SHAP computes the contributions of different parts of the time-domain sample $x$ to the prediction $\mathcal{M}(x)$ using \eqref{eq:SHAP_SHAP}, with attribution results corresponding to specific timings. While it offers clear interpretability for simple signals, it faces challenges in real-world high-noise scenarios, where fault features are often obscured by noise.

Compared to traditional time-domain SHAP, CS-SHAP introduces two main modifications: \textbf{\textit{1) Sample preprocessing}}: The CS transform $\mathcal{Z}$ preprocesses time-domain samples $x$ into CS samples, including the CS representation $\mathrm{CS}_x$ and phase information $\theta_x$. \textbf{\textit{2) Model integration}}: The inverse CS transform $\mathcal{Z}^{-1}$ is integrated with the end-to-end model $\mathcal{M}$, enabling CS samples as inputs without altering the original model architecture.

The proposed CS-SHAP has the following advantages: \textbf{\textit{1) Reliable interpretability label}}: The CS domain highlights fault information more effectively than the time domain, providing more accurate and reliable ground-truths for interpretability evaluation. \textbf{\textit{2) Clearer explanations}}: While the time domain can only reveal the timings of fault components, the CS domain simultaneously reveals both the carrier $f_c$ and modulation $f_m$ frequencies, which is critical for fault lacating and reasoning. \textbf{\textit{3) Noise robustness}}: The CS domain is less susceptible to noise interference than the time domain, making it more suitable for high-noise scenarios.

\subsection{Framework}

\begin{figure}[htbp]
    \centering
    \includegraphics[width=\textwidth]{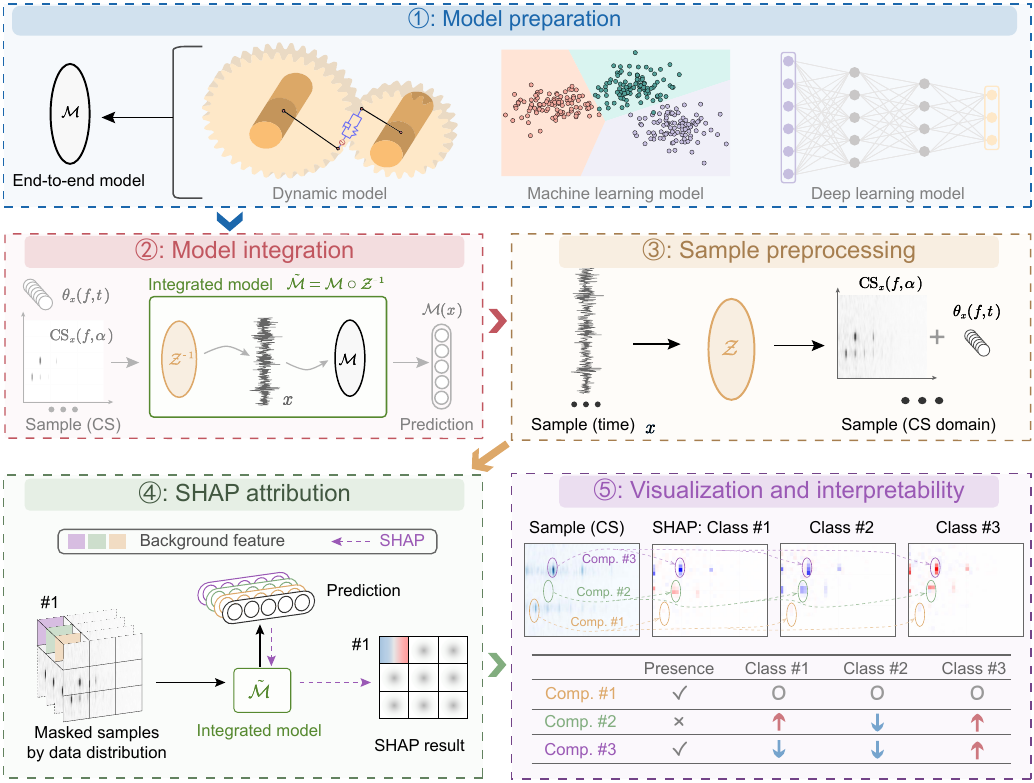}
    \caption{The flow chart of applying CS-SHAP for interpretability analysis.}
    \label{fig:framework}
\end{figure}

The workflow for applying CS-SHAP to end-to-end models for interpretability analysis is depicted in Fig.~\ref{fig:framework}. It comprises five key steps: model preparation, model integration, sample preprocessing, SHAP attribution, and visualization with interpretability analysis.

Firstly, users prepare an end-to-end model $\mathcal{M}$ based on their task requirements, which may derive from various approaches, including dynamics, machine learning, or deep learning. Secondly, the prepared model $\mathcal{M}$ is integrated with the CS transform $\mathcal{Z}$, creating an integrated model $\tilde{\mathcal{M}}$ capable of processing CS samples as input. Thirdly, existing time-domain samples $x$ are transformed into CS samples ($\mathrm{CS}_x$ and $\theta_x$) using the inverse CS transform $\mathcal{Z}^{-1}$. Next, SHAP analysis is conducted based on \eqref{eq:SHAP_SHAP}, iterating subsets and computing expectations relative to the data distribution to quantify the contribution of each part of the CS sample to the prediction $\mathcal{M}(x)$. Finally, the CS-SHAP results $\psi$ are visualized for interpretability analysis, with red indicating positive contributions and blue indicating negative contributions.

Notably, the contributions in CS-SHAP are influenced by two factors: the relevance of signal components to the current class and their presence. Specifically, the presence (absence) of fault components related to the current class results in a positive (negative) contribution, whereas the presence (absence) of fault components related to other classes leads to a negative (positive) contribution. 
As shown in the visualization part in Fig.~\ref{fig:framework}:
\textbf{Component \#1}, unrelated to all three classes, has zero contribution across all classes regardless of its presence. 
\textbf{Component \#2}, related to Class \#2, its absence contributes negatively to Class \#2 and positively to Classes \#1 and \#3. 
\textbf{Component \#3}, related to Class \#3, its presence contributes positively to Class \#3 but negatively to Classes \#1 and \#2.

\section{Experiments}

To validate the effectiveness of CS-SHAP, we applied it to three distinct datasets: a simulation dataset, the open-source Case Western Reserve University (CWRU) bearing dataset, and a private helical gearbox dataset. The simulation dataset, with fully known fault logic and ground-truth, serves as a reliable benchmark for evaluating interpretability results. The CWRU dataset, a widely used open-source benchmark, ensures the reproducibility of CS-SHAP using our open-source code. Finally, the private gearbox dataset, collected from a real industrial environment, highlights the practical applicability of CS-SHAP in real-world scenarios.

For the predictive model $\mathcal{M}$, we selected an end-to-end convolutional neural network (CNN) for analysis, with its architecture detailed in Table~\ref{tab:Exp-modelarch}. Similar to SHAP, CS-SHAP operates as a post-hoc interpretability method, making it model-agnostic. While the effects of different models will be discussed in Section~\ref{subsection:analysis-model}, we consistently utilize the CNN model for validation, ensuring that the assessment of CS-SHAP's interpretability performance remains unaffected.

\begin{table}[htbp]
        \centering
        \caption{The model architecture used in experiments.\label{tab:Exp-modelarch}}
        % \resizebox{\linewidth}{!}{
        \begin{threeparttable}
                \footnotesize
                \begin{tabular*}{\hsize}{@{\extracolsep{\fill}}lll}
                % \begin{tabular}{llll}
                        \toprule[1pt]
                        No. & Basic unit                                    & Output size     \\
                        \midrule[0.3pt]
                        -         & Input                                    & 1$\times$2000  \\
                        1         & Conv(8@7)\tnote{a}-BN-ReLU-MaxPool(2)       & 8$\times$997   \\
                        2$\sim$7:$\rightarrow i$  & {[}Conv($\textrm{2}^{i\textrm{+2}}$@3)-BN-ReLU-MaxPool(2){]}*6 & 512$\times$13  \\
                        9         & Conv(1024@3)-BN-ReLU-AdapMaxPool(1) & 1024$\times$1  \\
                        10        & Flatten-FC(256)-ReLU-FC(64)-ReLU-FC($K$) & $K$\tnote{b} \\
                        \bottomrule[1pt]
                \end{tabular*}

                \begin{tablenotes}
                        \smallskip
                        \footnotesize
                        \item[a] Conv(\textit{x}@\textit{y}): represents a convolutional layer with \textit{x} output channels and a kernel size of \textit{y}. Besides, the stride is 1, the padding is 0, and the input channels could be determined by the output size of the previous layer.
                        \item[b] $K$: represents the number of classes in the dataset.
                \end{tablenotes}
        \end{threeparttable}
        % }
\end{table}

For comparison, we selected several mainstream attribution methods: Grad-CAM, Time-SHAP (standard SHAP in the time domain), Freq-SHAP~\cite{herwigExplainingDeepNeural2023} (SHAP in the frequency domain), TF-SHAP~\cite{herwigExplainingDeepNeural2023} (SHAP in the time-frequency domain), and Env-SHAP~\cite{deckerDoesYourModel2023} (SHAP in the envelope domain).

\subsection{Simulation dataset}
We first define the periodic-impulse component $x_p$:
\begin{equation}
        \begin{split}
                x_p(f_m,f_c,t)\!=\!\sum_{k\in \mathbb{N}} e^{-\beta(t-k/f_m)} \sin \big( 2\pi f_c (t-\frac{k}{f_m})+\phi \big)
        \end{split}
\end{equation}
where, $f_m$ is the excitation frequency of the fault impulse, $f_c$ is the response frequency, $\beta=0.04$ is the damping, and $\phi \sim \mathcal{U} (0,2\pi)$ is the initial phase, where $\mathcal{U} (a,b)$ represents a uniform distribution with a lower bound $a$ and an upper bound $b$. The simulation signal can be represented as:
\begin{equation}
        x=\sum_{i=1}^{2} A x_p^i(f_m^i,f_c^i,t)+n(t)
\end{equation}
where, $A\sim\mathcal{U} (0.8,1)$ is the amplitude coefficient, and $n(t)$ represents Gaussian white noise with a signal-to-noise ratio (SNR) of 0.
% Samples from different classes have different $f_m$ and $f_i$.

\begin{figure}[htbp]
        \centering
        \includegraphics[width=\textwidth]{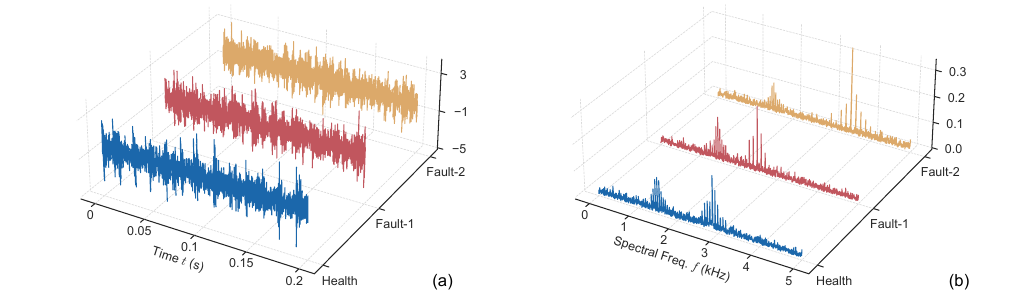}
        \caption{The representations of three classes in the simulation dataset. (a) Time domain. (b) Frequency domain.}
        \label{fig:4-1-signal}
\end{figure}

In this dataset, the sampling frequency is set to 10 kHz, and three fault classes are defined: health, Fault \#1, and Fault \#2, with their relationships to the components and corresponding parameters depicted in Table~\ref{tab:ExpSimulation-data}. Each fault class contains two periodic-impulse components. Specifically, Component $P_0$ is shared across all three classes and has zero contribution to any of them. In contrast, other components (i.e., $P_H$, $P_1$, $P_2$) are exclusive to a single class, contributing positively to their corresponding class while contributing negatively to others. The fault logic of this simulation dataset is fully known, facilitating the evaluation of interpretability that is challenging to achieve with other datasets due to the lack of ground-truth. For better understanding, the three fault classes in time domain and frequency domain are shown in Fig.~\ref{fig:4-1-signal}.

\begin{table}[htbp]
        \centering
        \caption{The parameters of the periodic-impulse components, along with their relationship with fault classes. \label{tab:ExpSimulation-data}}
        % \resizebox{\linewidth}{!}{
        \begin{threeparttable}
                \footnotesize
                \begin{tabular*}{\hsize}{@{\extracolsep{\fill}}cccccc}
                \toprule[1pt]
                Component & $f_c$ (kHz) & $f_m$ (Hz) & Health & Fault \#1 & Fault \#2 \\
                \midrule[0.3pt]
                $P_0$     & 1.5 & \multicolumn{1}{c|}{50} & \checkmark & \checkmark   & \checkmark   \\
                $P_H$   & $\mathcal{U}(1,4)$ & \multicolumn{1}{c|}{$\mathcal{U}(20,200)$} & \checkmark &          &          \\
                $P_1$ & 2.5 & \multicolumn{1}{c|}{100} &        & \checkmark   &          \\
                $P_2$ & 3.5 & \multicolumn{1}{c|}{125} &        &          & \checkmark   \\
                \bottomrule[1pt]
                \end{tabular*}

                % \begin{tablenotes}
                %         \smallskip
                %         \footnotesize
                %         \item[a] 1.
                % \end{tablenotes}
        \end{threeparttable}
        % }
\end{table}

This experiment could be regarded as a 3-class classification task. Each class contains 5000 samples, each with a length of 2000, normalized via mean-std normalization. Of these, 70\% are randomly selected for training, while the remaining 30\% are used for testing. Training parameters include 20 epochs, a batch size of 64, Adam optimizer, and a learning rate of 0.001 with a 0.99 decay per epoch. Ultimately, the trained CNN model achieved a test accuracy of 99.98\%.

\begin{figure}[htbp]
        \centering
        \includegraphics[width=\textwidth]{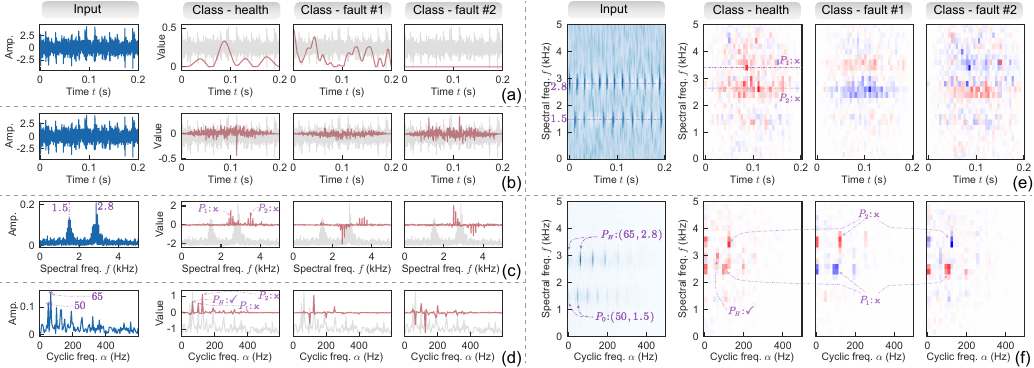}
        \caption{The domain representations of a health sample from the simulation dataset and its attribution results using various methods. (a) Grad-CAM. (b) Time-SHAP. (c) Freq-SHAP. (d) TF-SHAP. (e) Env-SHAP. (f) CS-SHAP.}
        \label{fig:4-1-health}
\end{figure}

\begin{figure}[htbp]
        \centering
        \includegraphics[width=\textwidth]{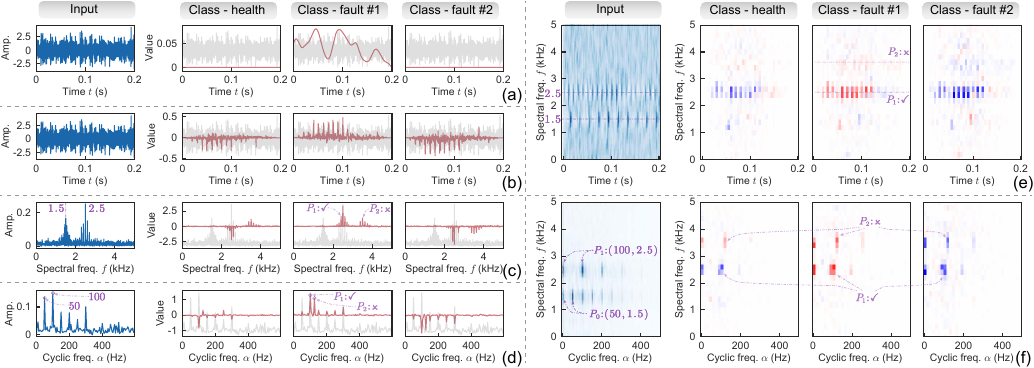}
        \caption{The domain representations of a Fault \#1 sample from the simulation dataset and its attribution results. (a) Grad-CAM. (b) Time-SHAP. (c) Freq-SHAP. (d) TF-SHAP. (e) Env-SHAP. (f) CS-SHAP.}
        \label{fig:4-1-fault1}
\end{figure}

\begin{figure}[htbp]
        \centering
        \includegraphics[width=\textwidth]{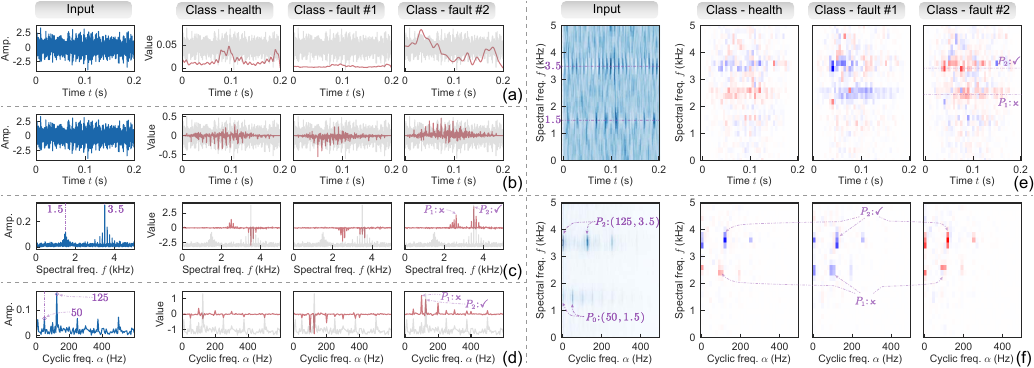}
        \caption{The domain representations of a Fault \#2 sample from the simulation dataset and its attribution results. (a) Grad-CAM. (b) Time-SHAP. (c) Freq-SHAP. (d) TF-SHAP. (e) Env-SHAP. (f) CS-SHAP.}
        \label{fig:4-1-fault2}
\end{figure}

% To evaluate the interpretability of CS-SHAP, we applied it to the trained CNN model and compared it with existing methods. 
The representations and attribution results of three class samples are shown in Fig.~\ref{fig:4-1-health}-\ref{fig:4-1-fault2}, respectively. First, we analyze the effect of different domains shown in the left side of Fig.~\ref{fig:4-1-health}-\ref{fig:4-1-fault2}. \textbf{\textit{1) Time domain}}: Impulse timings are obscured by noise, causing the lack of ground-truth. \textbf{\textit{2) Frequency domain}}: Periodic-impulse components appear as sidebands, revealing carrier $f_c$ and modulation $f_m$ frequencies. However, noise in real-world applications often masks $f_m$, limiting ground-truth to $f_c$. \textbf{\textit{3) Envelope spectrum domain}}: Periodic-impulse components are represented by modulation frequency $f_m$ and their harmonics, providing ground-truth for $f_m$. \textbf{\textit{4) Time-frequency domain}}: It reveals carrier frequency $f_c$ and impulse timings but is highly sensitive to noise, restricting ground-truth primarily to $f_c$. \textbf{\textit{5) CS domain}}: It clearly displays both carrier $f_c$ and modulation $f_m$ frequencies, including harmonics, and is robust to noise, offering comprehensive ground-truth for $f_c$ and $f_m$. In summary, the CS domain provides a complete representation of fault components, whereas other domains reveal partial information or are susceptible to noise interference.

The healthy sample comprises two components: the common $P_0$ and the random $P_{H}$. Its attribution results are shown in Fig.~\ref{fig:4-1-health}(a)-(f). \textbf{\textit{1) Grad-CAM \& Time-SHAP}}: Grad-CAM produces coarse and suboptimal results. Although Time-SHAP provides more refined outputs with certain impulse timings showing high contributions, it fails to offer meaningful insights. \textbf{\textit{2) Freq-SHAP}}:  The carrier frequencies $f_c$ of each component reveal that $P_0$ has no contribution on any class, while the prediction for the healthy class arises from the absence of $P_0$ and $P_1$.  \textbf{\textit{3) Env-SHAP}}: the modulation frequencies $f_m$ indicate that $P_0$ has no contribution, whereas the presence of $P_H$ and the absence of $P_1$ and $P_2$ positively contribute to the healthy class. \textbf{\textit{4) TF-SHAP}}: the results align with those from the Freq-SHAP, but are coarser due to variability in impulse timings. Based on their carrier frequencies, the absence of $P_1$ and $P_2$ positively contributes to the healthy class. \textbf{\textit{5) CS-SHAP}}: Based on their carrier frequencies and modulation frequencies, $P_0$ has no contribution, $P_H$'s presence contributes slightly positively to the healthy class and slightly negatively to the others, while the absence of $P_1$ and $P_2$ positively contributes to the healthy class and negatively contributes their corresponding classes. CS-SHAP not only has a comprehensive view of $f_m$ and $f_c$, but also aligns perfectly with the predefined fault logic.

The analysis approach for the remaining two classes follows the same method as that for the health sample, and we believe the attribution results in Fig.~\ref{fig:4-1-fault1}-\ref{fig:4-1-fault2} are straightforward for the reader to understand. For brevity, we only focus on the key points. Essentially, both Fault \#1 and Fault \#2 samples contain the common $P_0$ and the unique $P_1$ or $P_2$. Overall, because $P_0$ is shared across all classes, its presence contributes nothing to all the classes. $P_1$ corresponds to Fault \#1, so its presence contributes positively to Fault \#1 and negatively to the other classes, and vice versa. The effect of $P_2$ is similar to that of $P_1$, except that its positive contribution is associated with Fault \#2 instead. Fig.~\ref{fig:4-1-fault1}-\ref{fig:4-1-fault2} effectively validate the above logic, demonstrating the correctness of SHAP-based methods.

As for the explanations of different domains, both Grad-CAM and Time-SHAP perform poorly due to the variability in the impulse timings. The SHAP methods in other domains performs much better. Specifically, Freq-SHAP and TF-SHAP explain the contribution of the carrier frequency $f_c$, with the former being more refined and the latter providing a coarser granularity. Env-SHAP explains the contribution of the modulation frequency $f_m$. However, their interpretability results are partial, and only CS-SHAP can comprehensively explain the contributions of both carrier $f_c$ and modulation $f_m$ frequencies, providing a more complete and clearer explanation than others.

\subsection{Open-source CWRU bearing dataset}
\label{subsection:Exp-CWRU}

\begin{figure}[htbp]
        \centering
        \includegraphics[width=\textwidth]{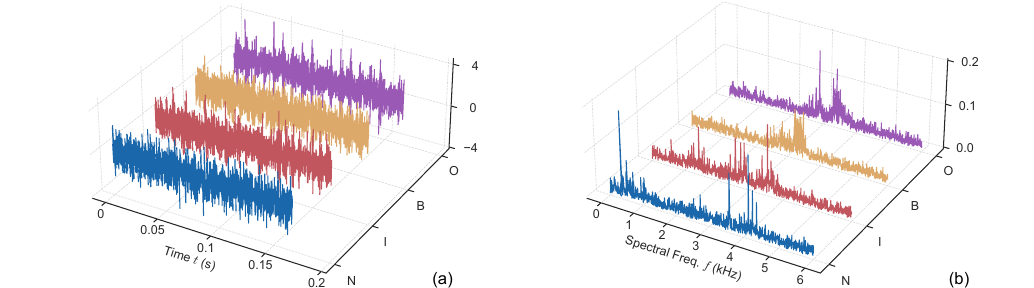}
        \caption{The representations of four classes in the CWRU bearing dataset. (a) Time domain. (b) Frequency domain.}
        \label{fig:4-2-signal}
\end{figure}

The CWRU dataset, a widely recognized benchmark in IFD research, is selected to validate the authenticity and reproducibility of CS-SHAP. This experiment chooses the operating condition of a 1 HP load at 1800 rpm. The corresponding bearing characteristic frequencies are listed in Table~\ref{tab:ExpCWRU-freq.}. Samples were collected from the drive end at a 12 kHz sampling rate, covering four classes: health (H), inner race fault (I), ball fault (B), and outer race fault (O), each with a defect size of 0.007 inches. Each class contains 119 samples of 2000 data points. For better understanding, the representations of the four classes in the time domain and frequency domain are shown in Fig.~\ref{fig:4-2-signal}, where the healthy sample prominently features an amplitude at 360 Hz ($nf_r$). The experimental and training settings were consistent with those of the simulation dataset. The end-to-end CNN model ultimately achieved a test accuracy of 100\%.

\begin{table}[htbp]
        \centering
        \caption{The characteristic frequencies of two datasets.\label{tab:ExpCWRU-freq.}}
        % \resizebox{\linewidth}{!}{
        \begin{threeparttable}
                \footnotesize
                \begin{tabular*}{\hsize}{@{\extracolsep{\fill}}ccc|ccc}
                        \toprule[1pt]
                        \multicolumn{3}{c}{CWRU bearing dataset\tnote{a} (Hz)}                                              & \multicolumn{3}{c}{Helical gearbox dataset\tnote{b} (Hz)}       \\
                        \midrule[0.3pt]
                        $f_r$ & $f_{\mathrm{BPFI}}$ & $f_{\mathrm{BPFO}}$ & $f_1$      & $f_2$        & $f_{\mathrm{mesh}}$     \\
                        30   & 162.45                                & 107.55                                & 30   & 7.683 & 630 \\
                        \bottomrule[1pt]
                \end{tabular*}

                \begin{tablenotes}
                        \smallskip
                        \footnotesize
                        \item[a] $n=12$: the number of rolling elements.  $f_r$: the rotation frequency. $f_{\mathrm{BPFI}}$: the ball pass frequency with inner race. $f_{\mathrm{BPFO}}$: the ball pass frequency with outer race. 
                        \item[b] $f_1$: the drive gear frequency. $f_2$: the driven gear frequency. $f_{\mathrm{mesh}}$: the meshing frequency.
                \end{tablenotes}
        \end{threeparttable}
        % }
\end{table}

\begin{figure}[htbp]
        \centering
        \includegraphics[width=\textwidth]{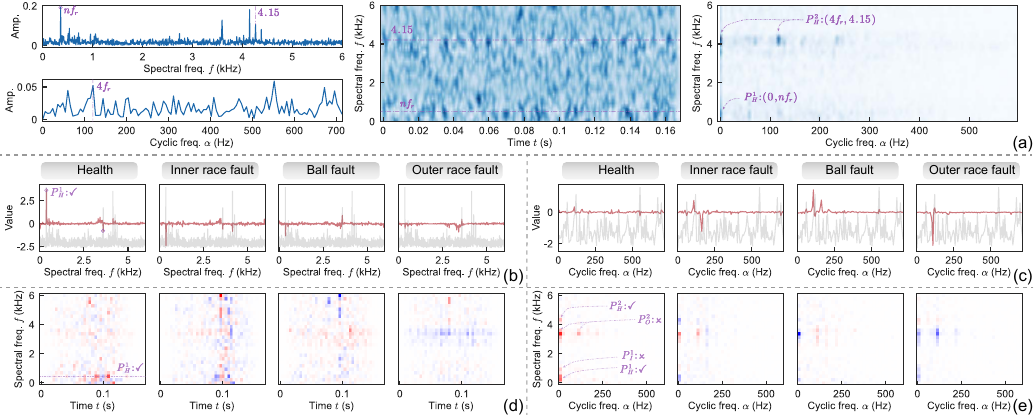}
        \caption{The domain representations of a health sample from the CWRU dataset and its attribution results. (a) Representations. (b) Freq-SHAP. (c) TF-SHAP. (d) Env-SHAP. (e) CS-SHAP.}
        \label{fig:4-2-health}
\end{figure}

\begin{figure}[htbp]
        \centering
        \includegraphics[width=\textwidth]{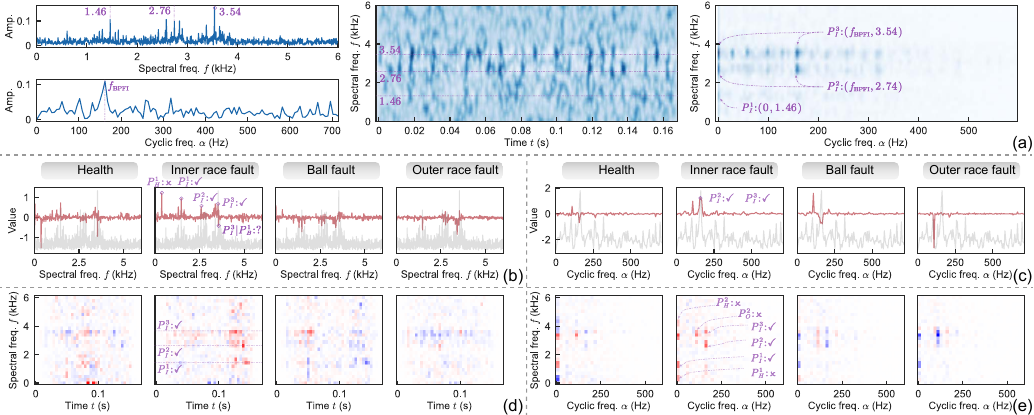}
        \caption{The domain representations of a inner race fault sample from the CWRU dataset and its attribution results. (a) Representations. (b) Freq-SHAP. (c) TF-SHAP. (d) Env-SHAP. (e) CS-SHAP.}
        \label{fig:4-2-IF}
\end{figure}

\begin{figure}[htbp]
        \centering
        \includegraphics[width=\textwidth]{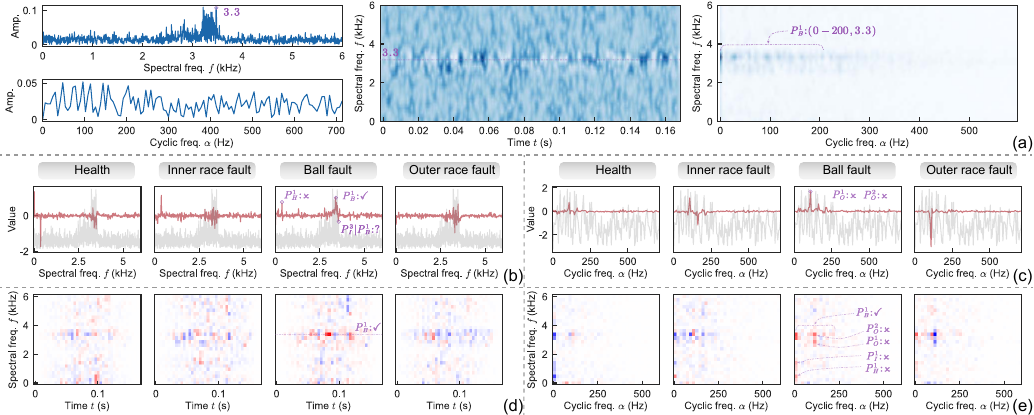}
        \caption{The domain representations of a ball fault sample from the CWRU dataset and its attribution results. (a) Representations. (b) Freq-SHAP. (c) TF-SHAP. (d) Env-SHAP. (e) CS-SHAP.}
        \label{fig:4-2-BF}
\end{figure}

\begin{figure}[htbp]
        \centering
        \includegraphics[width=\textwidth]{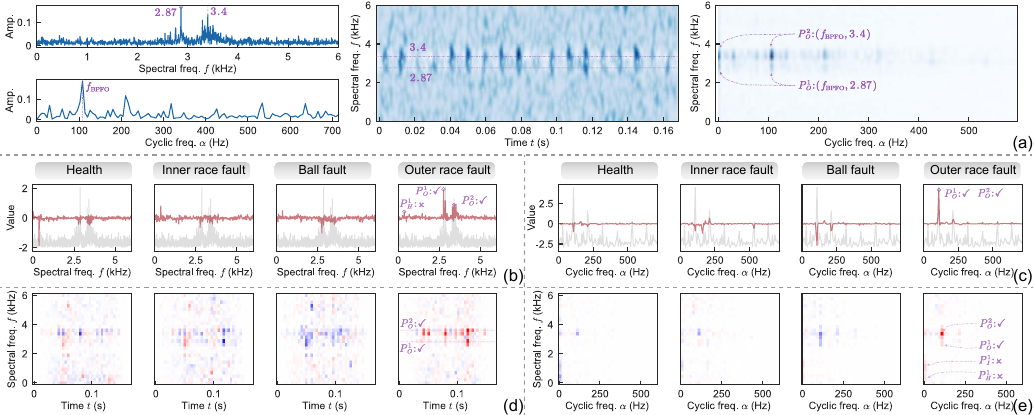}
        \caption{The domain representations of a outer race fault sample from the CWRU dataset and its attribution results. (a) Representations. (b) Freq-SHAP. (c) TF-SHAP. (d) Env-SHAP. (e) CS-SHAP.}
        \label{fig:4-2-OF}
\end{figure}

Unlike the simulation dataset where fault logic (ground truth) is explicitly defined,  the CWRU dataset requires the identification of fault characteristics for each class. For clarity, we denote modulated components as $P:(a,b)$, where $a$ (Hz) and $b$ (kHz) represent the carrier $f_c$ and modulation $f_m$ frequencies, respectively. The fault characteristics are as follows: The health sample in Fig.~\ref{fig:4-2-health}(a) contains a constant frequency component $P_H^1:(0,nf_r)$ and a modulated component $P_H^2:(4f_r,4.15)$. The inner race fault sample in Fig.~\ref{fig:4-2-IF}(a) includes $P_{I}^{1}:(0,1.46)$, $P_{I}^{2}:(f_{\mathrm{BPFI}},2.74)$, and $P_{I}^{3}:(f_{\mathrm{BPFI}},3.54)$. The ball fault sample in Fig.~\ref{fig:4-2-BF}(a) exhibits a broadband modulated component $P_{B}^{1}:(0-200,3.3)$. The outer race sample in Fig.~\ref{fig:4-2-OF}(a) contains $P_{O}^{1}:(f_{\mathrm{BPFO}},2.87)$ and $P_{O}^{2}:(f_{\mathrm{BPFO}},3.4)$. For simplicity, Grad-CAM and Time-SHAP are excluded due to their poor performance, and the remaining SHAP results are shown in Fig.~\ref{fig:4-2-health}–\ref{fig:4-2-OF}(b)-(e).

 As shown in Fig.~\ref{fig:4-2-health}(b), Freq-SHAP indicates that the contribution to health prediction $y_H$ comes from the presence of $P_H^1$. Env-SHAP in Fig.~\ref{fig:4-2-health}(c) reveals that no significant influence of modulated frequencies on $y_H$. TF-SHAP in Fig.~\ref{fig:4-2-health}(d) similarly attributes $y_H$ mainly to the presence of $P_H^1$, aligning with Freq-SHAP. However, CS-SHAP in Fig.~\ref{fig:4-2-health}(e), identifies that the $y_H$ contributions come from the presence of $P_H^1$ and $P_H^2$, as well as the the absence of $P_I^1$ and $P_O^2$, where $P_H^1$ and $P_O^2$ contribute more prominently. CS-SHAP further reveals that the constant frequency $f_m=0$ in $P_H^2$ contributes more significantly to $y_H$ than the modulated frequency $f_m=4f_r$, consistent with Env-SHAP. Moreover, CS-SHAP exhibits superior precision by capturing positive contributions resulting from the absence of other class components (i.e.,$P_I^1$ and $P_O^2$)—a phenomenon rarely identified by other SHAP methods. The analysis for Fig.~\ref{fig:4-2-IF}-\ref{fig:4-2-OF} are consistent with that of Fig.~\ref{fig:4-2-health} described above. With the detailed annotations provided in the figures, readers should find them easy to understand.

Notably, different fault components may share similar carrier frequencies $f_c$ or the same modulation frequency $f_m$, posing challenges for other SHAP methods to distinguish closely related components, potentially resulting in inaccuracies. For instance, $P_I^3$ and $P_B^1$ have similar $f_c$ around 3.3 kHz. In Fig.~\ref{fig:4-2-IF}(b), this similarity causes Freq-SHAP to incorrectly assign both positive and negative contributions to the 3.3 kHz frequency for the corresponding class $y_I$. This incorrect phenomenon is also observed in Fig.~\ref{fig:4-2-BF}(b). In contrast, CS-SHAP in Fig.~\ref{fig:4-2-IF}-\ref{fig:4-2-BF}(e) successfully differentiates $P_I^3$ and $P_B^1$ by leveraging their distinct $f_c$ and $f_m$, thereby avoiding this error. Likewise, $P_O^1$ and $P_O^2$ share the same modulation frequency $f_{\mathrm{BPIO}}$. As depicted in Fig.~\ref{fig:4-2-OF}(c), although Env-SHAP identifies $f_{\mathrm{BPIO}}$ as having a significant contribution, it cannot specify whether this contribution arises from $P_O^1$ or $P_O^2$. However, CS-SHAP in Fig.~\ref{fig:4-2-OF}(e) successfully separates these components, revealing that $P_O^2$ contributes more significantly than $P_O^1$. This result is consistent with the TF-SHAP result in Fig.~\ref{fig:4-2-OF}(f) and the CS representation in Fig.~\ref{fig:4-2-OF}(a).

In summary, the CWRU dataset further demonstrates the advantages of CS-SHAP over other SHAP methods. CS-SHAP provides more precise contribution results by considering not only the presence of fault components but also the absence of other fault components, which can also significantly contribute. Additionally, CS-SHAP more effectively differentiates fault components based on their carrier and modulation frequencies, leading to clearer contribution explanations and preventing potential attribution errors.

\subsection{Private helical gearbox dataset}

\begin{figure}[htbp]
        \centering
        \includegraphics[width=\textwidth]{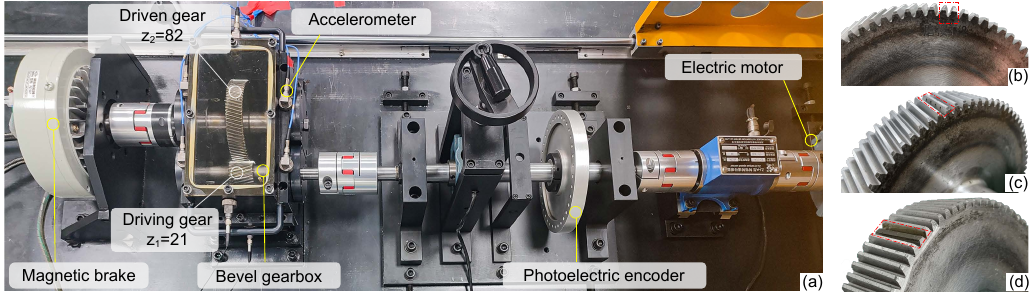}
        \caption{The experimental setup and fault types of the Helical gearbox. (a) The experimental setup. (b) W: Wear. (c) P: Pitting. (d) C: Tooth crack.}
        \label{fig:4-3-rig}
\end{figure}

\begin{figure}[htbp]
    \centering
    \includegraphics[width=\textwidth]{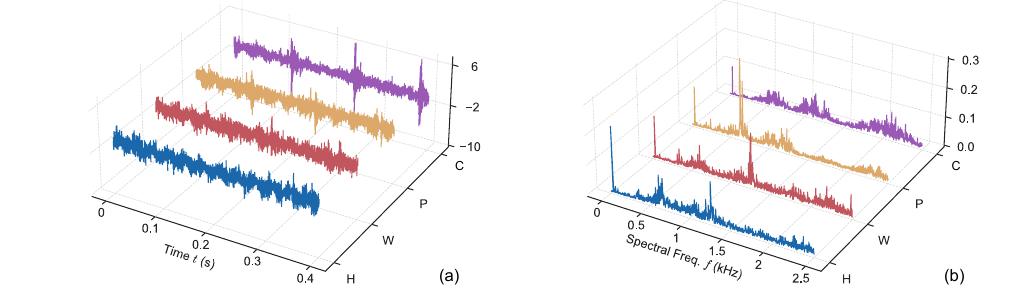}
    \caption{The representations of three classes in the  helical gearbox dataset. (a) Time domain. (b) Frequency domain.}
    \label{fig:4-3-signal}
\end{figure}

This dataset is collected from a private helical gearbox and is used to validate the practical effectiveness of CS-SHAP. The experimental setup and fault types are depicted in Fig.~\ref{fig:4-3-rig}. The motor speed is set to 1800 rpm, and the characteristic frequencies are summarized in Table~\ref{tab:ExpCWRU-freq.}. Data was acquired from the drive shaft end-shield at a 12 kHz sampling rate, comprising four classes: health (H), wear (W), pitting (P), and crack (C). Each class contains 76 samples of 2000 data points, and the representations of the four classes in the time domain and frequency domain are shown in Fig~\ref{fig:4-3-signal}. Experimental and training settings follow those of the simulation dataset, and the end-to-end CNN model achieves a test accuracy of 100\%.

\begin{figure}[htbp]
        \centering
        \includegraphics[width=\textwidth]{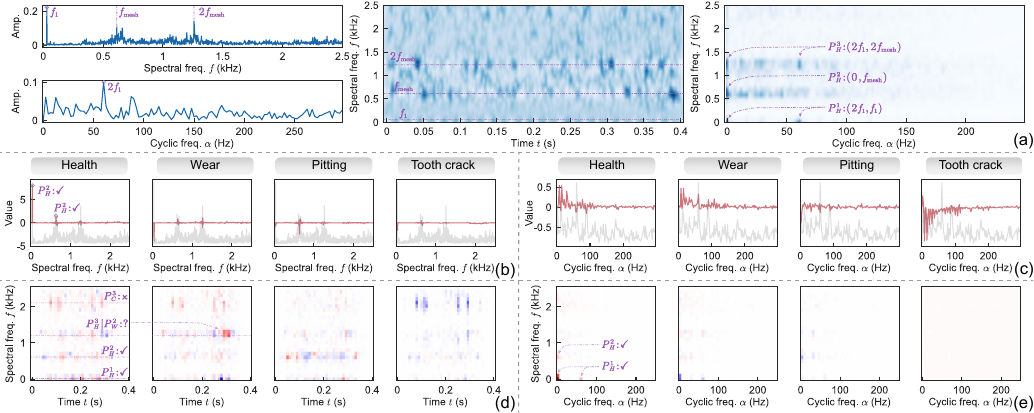}
        \caption{The domain representations of a health sample from the helical gearbox dataset and its attribution results. (a) Representations. (b) Freq-SHAP. (c) TF-SHAP. (d) Env-SHAP. (e) CS-SHAP.}
        \label{fig:4-3-health}
\end{figure}

\begin{figure}[htbp]
        \centering
        \includegraphics[width=\textwidth]{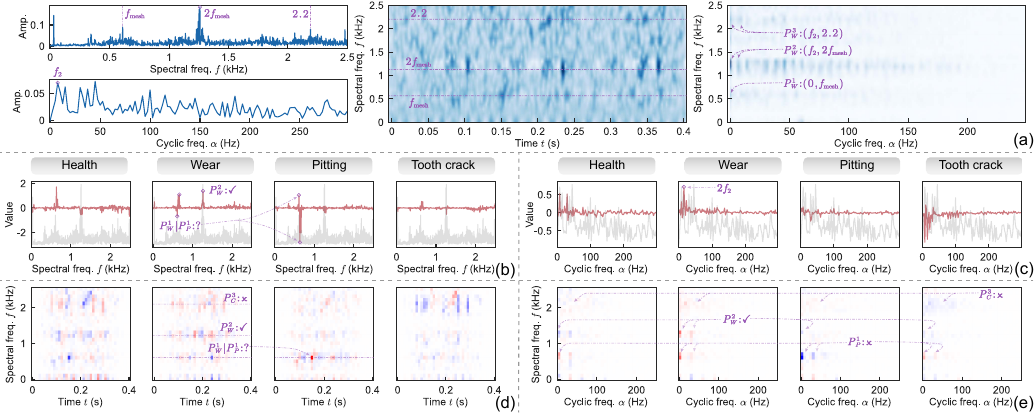}
        \caption{The domain representations of a wear fault sample from the helical gearbox dataset and its attribution results. (a) Representations. (b) Freq-SHAP. (c) TF-SHAP. (d) Env-SHAP. (e) CS-SHAP.}
        \label{fig:4-3-wear}
\end{figure}

\begin{figure}[htbp]
        \centering
        \includegraphics[width=\textwidth]{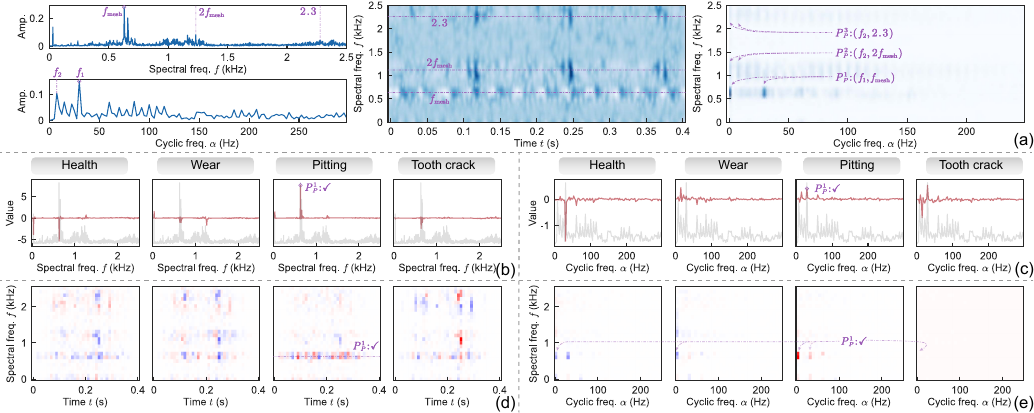}
        \caption{The domain representations of a pitting fault sample from the helical gearbox dataset and its attribution results. (a) Representations. (b) Freq-SHAP. (c) TF-SHAP. (d) Env-SHAP. (e) CS-SHAP.}
        \label{fig:4-3-pitting}
\end{figure}

\begin{figure}[htbp]
        \centering
        \includegraphics[width=\textwidth]{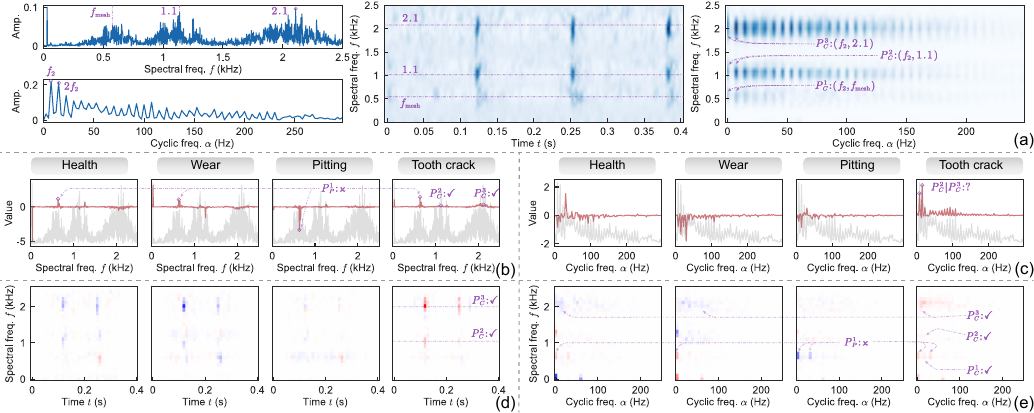}
        \caption{The domain representations of a tooth crack sample from the helical gearbox dataset and its attribution results. (a) Representations. (b) Freq-SHAP. (c) TF-SHAP. (d) Env-SHAP. (e) CS-SHAP.}
        \label{fig:4-3-crack}
\end{figure}

Similar to the CWRU dataset, ground-truth, i.e., the characteristics of each class, needs to be determined firstly. Specifically, the health sample in Fig.~\ref{fig:4-3-health}(a) includes $P_H^1:(2f_1,f_1)$, $P_H^2:(0,f_{\mathrm{mesh}})$, and $P_H^3:(2f_1,2f_{\mathrm{mesh}})$. The wear fault sample in Fig.~\ref{fig:4-3-wear}(a) contains $P_W^1:(0,f_{\mathrm{mesh}})$, $P_W^2:(f_2,2f_{\mathrm{mesh}})$, and $P_W^3:(f_2,2.2)$. The pitting fault sample in Fig.~\ref{fig:4-3-pitting}(a) exhibits $P_{P}^{1}:(f_1,f_{\mathrm{mesh}})$, $P_{P}^{2}:(f_2,f_{\mathrm{mesh}})$, and $P_{P}^{3}:(f_2,2.3)$. The tooth crack fault sample in Fig.~\ref{fig:4-3-crack}(a) contains $P_{C}^{1}:(f_2,f_{\mathrm{mesh}})$, $P_{C}^{2}:(f_2,1.1)$, and $P_{C}^{3}:(f_2,2.1)$. 

Unlike the CWRU dataset, the helical gearbox dataset has greater challenges for interpretability due to the prevalence of components with shared carrier frequencies $f_c$ or modulation frequencies $f_m$. As for the carrier frequency $f_c$, $P_W^2$ and $P_P^2$ share $f_c=2f_{\mathrm{mesh}}$. As shown in Fig.~\ref{fig:4-3-wear}(b), the presence of $2f_{\mathrm{mesh}}$ causes Freq-SHAP to attribute both positive and negative contributions to both $y_W$ and $y_P$, leading to contradictory results. In contrast, CS-SHAP in Fig.~\ref{fig:4-3-wear}(e) effectively differentiates between $P_W^2$ and $P_P^2$, assigning entirely positive contributions to $y_W$ and entirely negative contributions to $y_P$, thereby resolving the ambiguity.

Regarding the modulation frequency $f_m$, $P_C^2$ and $P_C^3$ share $f_m=f_2$, with both exhibiting significant energy in Fig.~\ref{fig:4-3-crack}(a). Env-SHAP in Fig.~\ref{fig:4-3-crack}(c) indicates that $f_2$ contribute significantly to $y_C$, but fails to distinguish whether these contributions originate from $P_C^2$ or $P_C^3$, requiring carrier frequency $f_c$ for clarification. Freq-SHAP in Fig.~\ref{fig:4-3-crack}(b) loses its attribution capability, as the $f_c$ of $P_C^2$ and $P_C^3$ are assigned nearly negligible contributions, which is unreasonable and highlights the limitations of such methods. Conversely, TF-SHAP in Fig.~\ref{fig:4-3-crack}(d) identifies the contribution of $P_C^3$ as greater than $P_C^2$. Similarly, CS-SHAP in Fig.~\ref{fig:4-3-crack}(e) shows that the contribution of $P_C^3$ is the most significant, roughly aligning with the results in Fig.~\ref{fig:4-3-crack}(d).

In summary, CS-SHAP provides a comprehensive explanation by both carrier frequency $f_c$ and modulation frequency $f_m$. On the one hand, it effectively separates close components, resulting in clearer explanation. On the other hand, its two-dimensional attribution (i.e., $f_c$ and $f_m$) aligns more closely with fault characteristics, enabling more accurate explanations. This approach avoids pitfalls, such as those seen in Fig.~\ref{fig:4-3-crack}(b), where most components are wrongly assigned negligible contributions.

\section{Analysis}

As indicated in \eqref{eq:SHAP_SHAP}, SHAP explanations depend on two primary factors: the model $\mathcal{M}$ and the data $x$. Thus, we analyze the CS-SHAP attribution results under  different models and noise intensities. All analyses are conducted on the CWRU dataset, following the same experimental settings described in Section~\ref{subsection:Exp-CWRU}.

\subsection{CS-SHAP with different models}
\label{subsection:analysis-model}

\begin{figure}[htbp]
        \centering
        \includegraphics[width=\textwidth]{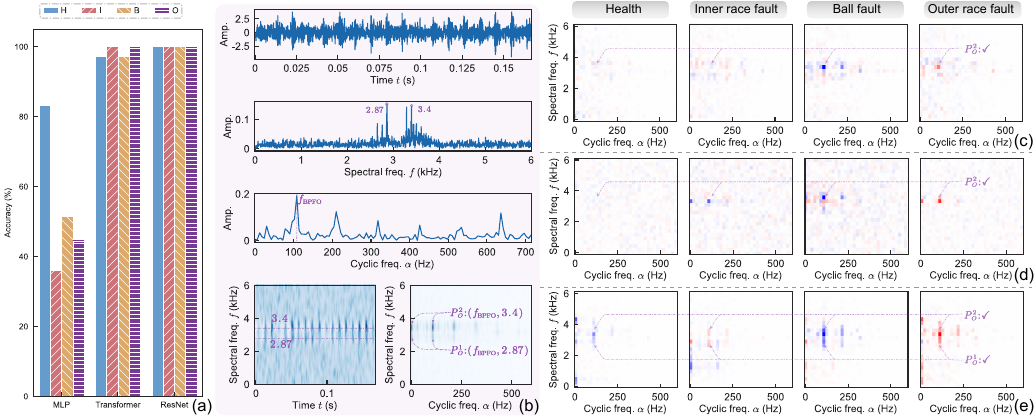}
        \caption{The class-wise test accuracies, domain representations, and CS-SHAP results of three models under the CWRU dataset. (a) Class-wise test accuracies. (b) Domain representations of the outer race fault sample for analysis. (c) CS-SHAP result of MLP with $y_O$ = 69.0\%. (d) CS-SHAP result of Transformer with $y_O$ = 100\%. (e) CS-SHAP result of ResNet with $y_O$ = 100\%.}
        \label{fig:5-1}
\end{figure}

We selected three representative models $\mathcal{M}$: Multilayer perceptron (MLP), Transformer, and ResNet. Their class-wise test accuracies are shown in Fig.~\ref{fig:5-1}(a). 
% The MLP's classification accuracy is around 50\%, with the accuracy for the healthy class increasing to 80\%, likely due to its distinct rotational frequency characteristic $nf_r$. Both the Transformer and ResNet exhibit excellent performance, achieving near 100\% classification accuracy. 
For CS-SHAP analysis, the same outer race fault sample is selected for all three models. Its multi-domain representations, shown in Fig.~\ref{fig:5-1}(b), reveal two components, $P_O^1$ and $P_O^2$.

The CS-SHAP results for the three models are shown in Fig.~\ref{fig:5-1}(c)-(e). the MLP's weak classification capability significantly impulses the CS-SHAP results, with the contribution of $P_O^1$ nearly zero and $P_O^2$ present but not prominent. In Fig.~\ref{fig:5-1}(d), the Transformer demonstrates improved classification performance, leading to a notable increase in $P_O^2$'s contribution; however, $P_O^1$'s contribution remains weak. Furthermore, slight amplitudes observed in blank regions suggest potential instability in the Transformer's results, likely due to the attention mechanism. In Fig.~\ref{fig:5-1}(e), the ResNet achieves the best performance, with both $P_O^1$ and $P_O^2$ displaying clear contributions, and $P_O^1 < P_O^2$. Considering the CS representation in Fig.~\ref{fig:5-1}(b), this result is quite reasonable.

In summary, the results of CS-SHAP are indeed influenced by the model $\mathcal{M}$. On one hand, weak classification capabilities lead to less distinct predictions, making it challenging for CS-SHAP to effectively capture contributions. On the other hand, the model type plays a critical role. For example, Transformers, which excel at capturing token relationships through the attention mechanism, may exhibit unstable results compared to CNNs. Nevertheless, CS-SHAP consistently provides generally accurate interpretability results across different models $\mathcal{M}$, demonstrating its broad applicability and potential as a universal interpretability algorithm for IFD models.

\subsection{CS-SHAP with different noise intensities}

\begin{figure}[htbp]
        \centering
        \includegraphics[width=\textwidth]{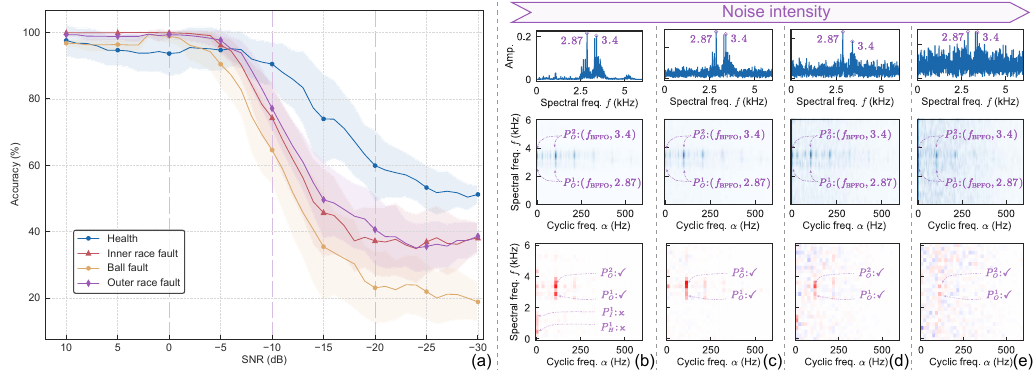}
        \caption{The class-wise test accuracies with different noise intensities under CWRU dataset, alongside frequency spectrum, CS representations, and CS-SHAP results of the outer race fault sample for prediction $y_O$. (a) Class-wise test accuracies. (b) Noise-free case with $y_O$ = 100\%. (c) SNR = 10 case with $y_O$ = 100\%. (d) SNR = 0 case with $y_O$ = 80.22\%. (e) SNR = -10 case with $y_O$ = 33.66\%.}
        \label{fig:5-2}
\end{figure}

Based on the CNN presented in Table~\ref{tab:Exp-modelarch}, The test accuracies under different noise intensities on the CWRU dataset are shown in Fig.~\ref{fig:5-2}(a). For CS-SHAP analysis, an outer race fault sample is selected, and its frequency spectrum, CS representation, and CS-SHAP results are shown in Fig.~\ref{fig:5-2}(b)-(e), respectively.

Firstly, under high noise conditions, $P_O^1$ and $P_O^2$ are more clearly distinguishable in the CS representation compared to the frequency spectrum, highlighting the superior feature extraction capability of CS representation. Furthermore, as noise intensity increases, the prediction difficulty rises significantly, leading to progressively less distinct CS-SHAP results. Nevertheless, CS-SHAP consistently reveals the contributions of  $P_O^1$ and $P_O^2$ to $y_O$. Even at SNR = -10 in Fig.~\ref{fig:5-2}(e), the contributions of $P_O^1$ and $P_O^2$ are weakened but still discernible.

In conclusion, owing to the strong feature extraction capability of CS representation, CS-SHAP demonstrates outstanding robustness to noise. Even under low classification probabilities, it still can calculate the contributions of key components, as shown in Fig.~\ref{fig:5-2}(e).

\section{Conclusion}
The form of explanation remains a significant challenge for existing IFD interpretability methods. To address this, we propose CS-SHAP, which extends SHAP to the CS domain, where fault features are more distinct and interpretable. Unlike existing SHAP extension approaches, CS-SHAP evaluates the contributions of components from both carrier frequency and modulation frequency perspectives, aligning closely with fault mechanisms. This dual-dimensional analysis allows CS-SHAP to effectively distinguish close fault components and excel in high-noise and multi-class scenarios. Comprehensive validation on simulation, CWRU, and private datasets demonstrates the superior interpretability, authenticity, and practicality of CS-SHAP. By publicly releasing the code, we aim to establish CS-SHAP as a benchmark post-hoc interpretability method for IFD and beyond.

Nevertheless, several challenges remain in advancing the interpretability of IFD. First, the validation of interpretability often lacks objective ground-truths, relying on subjective human judgment, which can compromise the reliability of explanations. Second, the interpretability inherently involves human involvement, necessitating the development of objective and quantifiable metrics to evaluate explanations comprehensively. Finally, while our work enhances explanation forms, the high-dimensional nature of vibration signals poses a significant challenge to the computational efficiency of interpretability methods, warranting further exploration.

\section*{Acknowledge}
This work was supported by the National Key Research and Development Program of China under Grant 2024YFB3409101, National Natural Science Foundation of China under Grant 12272219, and Grant 12121002, Opening Found of State Key Laboratory of Mechanical Transmission for Advanced Equipment under Grant SKLMT-MSKFKT-202303.

% \appendix
% \section{My Appendix}
% Appendix sections are coded under \verb+\appendix+.

% \verb+\printcredits+ command is used after appendix sections to list 
% author credit taxonomy contribution roles tagged using \verb+\credit+ 
% in frontmatter.

% %% Loading bibliography style file
% \bibliographystyle{model1-num-names}
% \bibliographystyle{cas-model2-names}
\bibliographystyle{elsarticle-num}

% % Loading bibliography database
\bibliography{reference.bib}

\end{document}